\documentclass[10pt,twocolumn,letterpaper]{article}

\usepackage{cvpr}              

\usepackage{graphicx}
\usepackage{amsmath}
\usepackage{amssymb}
\usepackage{booktabs}
\usepackage{nicefrac}
\usepackage{multirow}
\renewcommand{\paragraph}[1]{\vspace{1.2mm}\noindent\textbf{#1}}

\usepackage[pagebackref,breaklinks,colorlinks]{hyperref}

\usepackage[capitalize]{cleveref}
\crefname{section}{Sec.}{Secs.}
\Crefname{section}{Section}{Sections}
\Crefname{table}{Table}{Tables}
\crefname{table}{Tab.}{Tabs.}

\def\cvprPaperID{*****} 
\def\confName{CVPR}
\def\confYear{2023}

\begin{document}
\title{M2Former: Multi-Scale Patch Selection for Fine-Grained Visual Recognition}

\author{Jiyong Moon$^1$ \qquad Junseok Lee$^1$ \qquad Yunju Lee$^1$ \qquad Seongsik Park$^1$\thanks{corresponding author} \vspace{.5em} \\
$^1$Dongguk University \vspace{.5em} \\
{\tt\small \{asdwldyd, 2021126776, 2022126715, s.park\}@dgu.ac.kr}
}
\maketitle

\begin{abstract}
Recently, vision Transformers (ViTs) have been actively applied to fine-grained visual recognition (FGVR). ViT can effectively model the interdependencies between patch-divided object regions through an inherent self-attention mechanism. In addition, patch selection is used with ViT to remove redundant patch information and highlight the most discriminative object patches. However, existing ViT-based FGVR models are limited to single-scale processing, and their fixed receptive fields hinder representational richness and exacerbate vulnerability to scale variability. Therefore, we propose multi-scale patch selection (MSPS) to improve the multi-scale capabilities of existing ViT-based models. Specifically, MSPS selects salient patches of different scales at different stages of a multi-scale vision Transformer (MS-ViT). In addition, we introduce class token transfer (CTT) and multi-scale cross-attention (MSCA) to model cross-scale interactions between selected multi-scale patches and fully reflect them in model decisions. Compared to previous single-scale patch selection (SSPS), our proposed MSPS encourages richer object representations based on feature hierarchy and consistently improves performance from small-sized to large-sized objects. As a result, we propose M2Former, which outperforms CNN-/ViT-based models on several widely used FGVR benchmarks.
\end{abstract}

\section{Introduction}
\label{sec:intro}
\begin{figure}[t]
  \centering
   \includegraphics[width=1.0\linewidth]{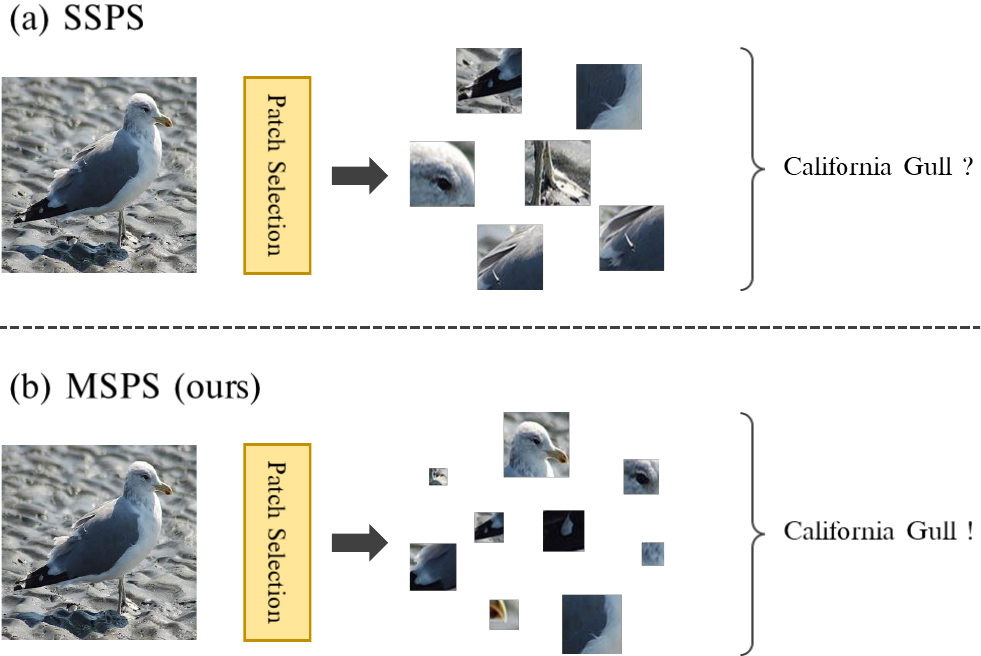}

   \caption{Comparison between previous single-scale patch selection (SSPS) and our multi-scale patch selection (MSPS). (a) SSPS extracts salient image patches of the same size, and the limited receptive field causes suboptimal object representation and vulnerability to scale variance. (b) On the other hand, our MSPS extracts salient patches in multi-scale. This encourages rich representations of objects, from deep semantic information in large-sized patches to fine-grained details in small-sized patches. In addition, the flexibility of multi-scale patches is useful for handling extremely large or small objects through multiple receptive fields.}
   \label{fig1}
\end{figure}
Despite recent rapid advances, fine-grained visual recognition (FGVR) is still one of the non-trivial tasks in computer vision community. Unlike conventional recognition tasks, FGVR aims to predict subordinate categories of a given object, \textit{e.g.}, subcategories of birds~\cite{DATA1, DATA2}, flowers~\cite{DATA3, DATA4} and cars~\cite{DATA5, DATA6}. It is a highly challenging task due to inherently subtle inter-class differences caused by similar subordinate categories and large intra-class variations caused by object pose, scale, or deformation.

The most common solution for FGVR is to decompose the target object into multiple local parts~\cite{FGPA1, FGPA2, FGPA3, FGWS1, FGWS2, FGWS3}. Due to subtle differences between fine-grained categories mostly resides in the unique properties of object parts~\cite{FGPA4}, decomposed local parts provide more discriminative clues of the target object. For example, a given bird object can be decomposed into its beak, wing, and head parts. At this time, \textit{Glaucous Winged Gull} and \textit{California Gull} can be distinguished by comparing their corresponding object parts. Early approaches of these \textit{part-based} methods finds discriminative local parts using manual part annotations~\cite{FGPA1, FGPA2, FGPA3, FGPA4}. However, curating manual annotations for all possible object parts is labor-intensive and carries the risk of human-error~\cite{FGET1}. Therefore, research focus has consequently shifted to a weakly-supervised manner~\cite{FGWS1, FGWS2, FGWS3, FGWS10, FGWS11, FGWS12}. They use additional tricks such as attention mechanisms~\cite{FGWS3, FGWS11, FGWS15} or region proposal networks (RPN)~\cite{FGWS10, FGWS12, FGWS14} to estimate local parts with only category-level labels. However, the part proposal process greatly increases the overall computational cost. Additionally, they tend not to deeply consider the interactions between estimated local parts that are essential for accurate recognition~\cite{FGTR1}.

Recently, vision Transformers (ViTs)~\cite{ETC1} are being actively applied to FGVR~\cite{FGTR1, FGTR2, FGTR3, FGTR4, FGTR5, FGTR6, FGTR7, FGTR8, FGTR9}. Relying exclusively on the Transformer~\cite{ETC2} architecture, ViT has shown competitive image classification performance at large-scale. Similar to token sequences in NLP, ViT embeds the input images into fixed-size image patches, and the patches pass through multiple Transformer encoder blocks. ViT's patch-by-patch processing is highly suitable for FGVR because each image patch can be considered as a local part. This means that the cumbersome part proposal process is no longer necessary. Additionally, the self-attention mechanism~\cite{ETC2} inherent in each encoder block facilitates modeling of global interactions between patch-divided local parts. ViT-based FGVR models use patch selection to further boost the performance~\cite{FGTR1, FGTR2, FGTR3, FGTR4}. Because ViT deals with all patch-divided image regions equally, many irrelevant patches may lead to inaccurate recognition. Similar to part proposals, patch selection selects the most salient patches from a set of generated image patches based on the computed importance ranking, \textit{i.e.}, accumulated attention weights~\cite{ETC3, FGTR3}. As a result, redundant patch information is filtered out and only selected salient patches are considered for final decision.

However, the existing ViT-based FGVR methods suffer from their single-scale limitations. ViT uses fixed-size image patches throughout the entire network, which ensures that the receptive field remains the same across all layers, and it prevents the ViT from obtaining multi-scale feature representations~\cite{TRMS2, TRMS4, TRMS5}. On the other hand, convolutional neural networks (CNNs) are suitable for multi-scale feature representations thanks to their staged architecture, where feature resolution decreases as layer depth increases~\cite{MULT3, ETC4, ETC5, MULT14, MULT15}. In early stages, spatial details of an object are encoded on high-resolution feature maps, and as the stages deepens, the receptive field expands with decreasing feature resolution, and higher-order semantic patterns are encoded into low-resolution feature maps. Multi-scale features are important for most vision tasks, especially pixel-level dense predictions tasks, \textit{e.g.}, object detection~\cite{MULT4, MULT5, MULT6}, and segmentation~\cite{MULT7, MULT8, MULT9}. In the same context, single-scale processing can cause two failure cases in FGVR, which lead to suboptimal recognition performance. (i) First, it is vulnerable to scale changes of fine-grained objects~\cite{TRMS5, ETC5, MULT4}. Fixed patch size may be insufficient to capture very subtle features of small-scale objects due to too coarse patches, and conversely, discriminative features may be over-decomposed for large-scale objects due to too finely split patches. (ii) Second, single-scale processing limits representational richness for objects~\cite{TRMS2, MULT5}. Compared to CNN that explores rich feature hierarchies from multi-scale features, ViT considers only monotonic single-scale features due to its fixed receptive field.

In this paper, we improve existing ViT-based FGVR methods by enhancing multi-scale capabilities. One simple solution is to use the recent multi-scale vision Transformers (MS-ViT)~\cite{TRMS1, TRMS2, TRMS3, TRMS4, TRMS5, TRMS6, TRMS7, TRMS8, TRMS9}. In fact, we can achieve satisfactory results simply by using MS-ViT. However, we further boost the performance by adapting patch selection to MS-ViT. Specifically, we propose a multi-scale patch selection (MSPS) that extends the previous single-scale patch selection (SSPS)~\cite{FGTR1, FGTR2, FGTR3, FGTR4} to multi-scale. MSPS select salient patches of different scales from different stages of the MS-ViT backbone. As shown in Fig.~\ref{fig1}, multi-scale salient patches selected through MSPS include both large-scale patches that capture object semantics and small-scale patches that capture fine-grained details. Compared to single-scale patches in SSPS, feature hierarchies in multi-scale patches provide richer representations of objects, which leads to better recognition performance. In addition, the flexibility of multi-scale patches is useful for handling extremely large/small objects through multiple receptive fields. 

However, we argue that patch selection alone cannot fully explain the object, and consideration is required of \textit{how to model interactions between selected patches and effectively reflect them in the final decision}. It is more complicated than the case considering only single-scale patches. Therefore, we introduce class token transfer (CTT) and multi-scale cross-attention (MSCA) to effectively deal with selected multi-scale patches. First, CTT aggregates the multi-scale patch information by transferring the global \texttt{CLS} token to each stage. Each stage-specific patch information is shared through transferred global \texttt{CLS} tokens, which generate richer network-level representations. In addition, we propose MSCA to model direct interactions between selected multi-scale patches. In the MSCA block, cross-scale interactions in both spatial and channel dimensions are computed for selected patches of all stages. Finally, our \textit{multi-scale vision Transformer with multi-scale patch selection} (\textbf{M2Former}) obtains improved FGVR performance over other ViT-based SSPS models, as well as CNN-based models.

Our main contributions can be summarized as follows:

\begin{itemize}
\item We propose MSPS that further boosts the multi-scale capabilities of MS-ViT. Compared to SSPS, MSPS generates richer representations of fine-grained objects with feature hierarchies, and obtains flexibility for scale changes with multiple receptive fields.
\item We propose CTT that effectively shares the selected multi-scale patch information. Stage-specific patch information is shared through transferred global \texttt{CLS} tokens to generate enhanced network-level representations.
\item We design multi-scale cross-attention (MSCA) block to capture the direct interactions of selected multi-scale patches. In the MSCA block, the spatial-/channel-wise cross-scale interdependencies can be captured.
\item Extensive experimental results on widely used FGVR benchmarks show the superiority of our M2Former over conventional methods. In short, our M2Former achieves an accuracy of 92.4\% on Caltech-UCSD Birds (CUB)~\cite{DATA1}, and 91.1\% on NABirds~\cite{DATA7}.
\end{itemize}
%
%
\section{Related Work}
\label{sec:related}
\subsection{Part-based FGVR}
\label{subsec:fgir}
A number of methods have been proposed to classify subordinate object categories~\cite{FGPA1, FGPA2, FGPA3, FGPA4, FGWS10, FGWS11, FGWS12, FGWS13, FGET1, FGET2, FGET3, FGET4, FGET5, FGET6, FGET17}. Among them, part-based methods decompose target objects into multiple local parts to capture more discriminative features~\cite{FGPA1, FGPA2, FGPA3, FGPA4, FGWS10, FGWS11, FGWS13}. Encoded local representations can be used either in conjunction with image-level representations or by themselves for recognition. This entails the use of a detection branch to generate discriminative part proposals from the input image either before or in parallel with the classification layer. Early works leverage manual part annotations for a fully-supervised train of detection branches~\cite{FGPA1, FGPA2, FGPA3, FGPA4}. However, curating large-scale part annotations is labor-intensive and highly susceptible to human error. Thus, recent part-based methods localize informative regions in a weakly-supervised way using only category-level labels~\cite{FGWS3, FGWS4, FGWS6, FGWS7, FGWS9, FGWS10, FGWS11, FGWS12, FGWS13, FGWS14, FGWS15}. TASN~\cite{FGWS2} introduces trilinear attention sampling to generate detail-preserved views based on feature channels. P2P-Net~\cite{FGWS1} and DF-GMM~\cite{FGWS5} use a Feature Pyramid Network (FPN)~\cite{ETC5} to generate local part proposals from convolutional feature maps. RA-CNN~\cite{FGWS13} iteratively zoom in local discriminative regions and reinforce multi-scale feature learning with inter-scale ranking loss. InSP~\cite{FGWS8} extracts potential part regions from the attention maps of two different images and applies content swapping to learn fine-grained local structures. Those part-based methods are good at locating discriminative parts, but they are ambiguous at capturing interactions between estimated local parts. Additionally, explicitly generating region proposals incurs a large computational burden, and performance can degrade significantly if the generated proposals are inaccurate. Thus, we focus on ViT-based FGVR methods that can avoid cumbersome region proposals.
\subsection{ViT-based FGVR}
\label{subsec:vit}
Transformer~\cite{ETC2} is originally introduced in the natural language processing (NLP) community, and recently shown great potential in the field of computer vision. Specifically, vision Transformer (ViT)~\cite{ETC1} has recorded remarkable performance in image classification on large-scale datasets relying entirely on a pure Transformer encoder architecture. ViT splits images into non-overlapping patches and then applies multiple Transformer encoder blocks, consisting of multi-head self-attention module (MSA) and feed-forward networks (FFN). Recently, ViTs are being actively applied to FGVR~\cite{FGTR1, FGTR2, FGTR3, FGTR4, FGTR5, FGTR6, FGTR7, FGTR8, FGTR9}. ViT is highly suitable for FGVR in that (i) patch-by-patch processing can effectively replace part proposal generation, and (ii) it is easy to model global interactions between patch-divided local parts through the inherent self-attention mechanism. In most cases, patch selection, which selects discriminative patches from a initial patch sequence, is used with ViT-based models. TransFG~\cite{FGTR1} selects the most discriminative patches based on the attention score at the last encoder block to consider only the important image regions. Similarly, DCAL~\cite{FGTR2} conducts patch selection based on attention scores to highlight the interaction of high-response image regions. FFVT~\cite{FGTR4} extends this selection process to all layers of ViT to utilize layer-wise patch information. RAMS-Trans~\cite{FGTR3} enhances local representations by recurrently zooming-in salient object regions obtained by patch selection. For the same purpose, we use patch selection to strengthen the decisiveness of the network. However, we extend the existing patch selection from single-scale to multi-scale to improve representational richness and obtain flexibility for scale changes.
\subsection{Multi-Scale Processing}
\label{subsec:msp}
\begin{figure*}[ht]
  \centering
   \includegraphics[width=1.0\linewidth]{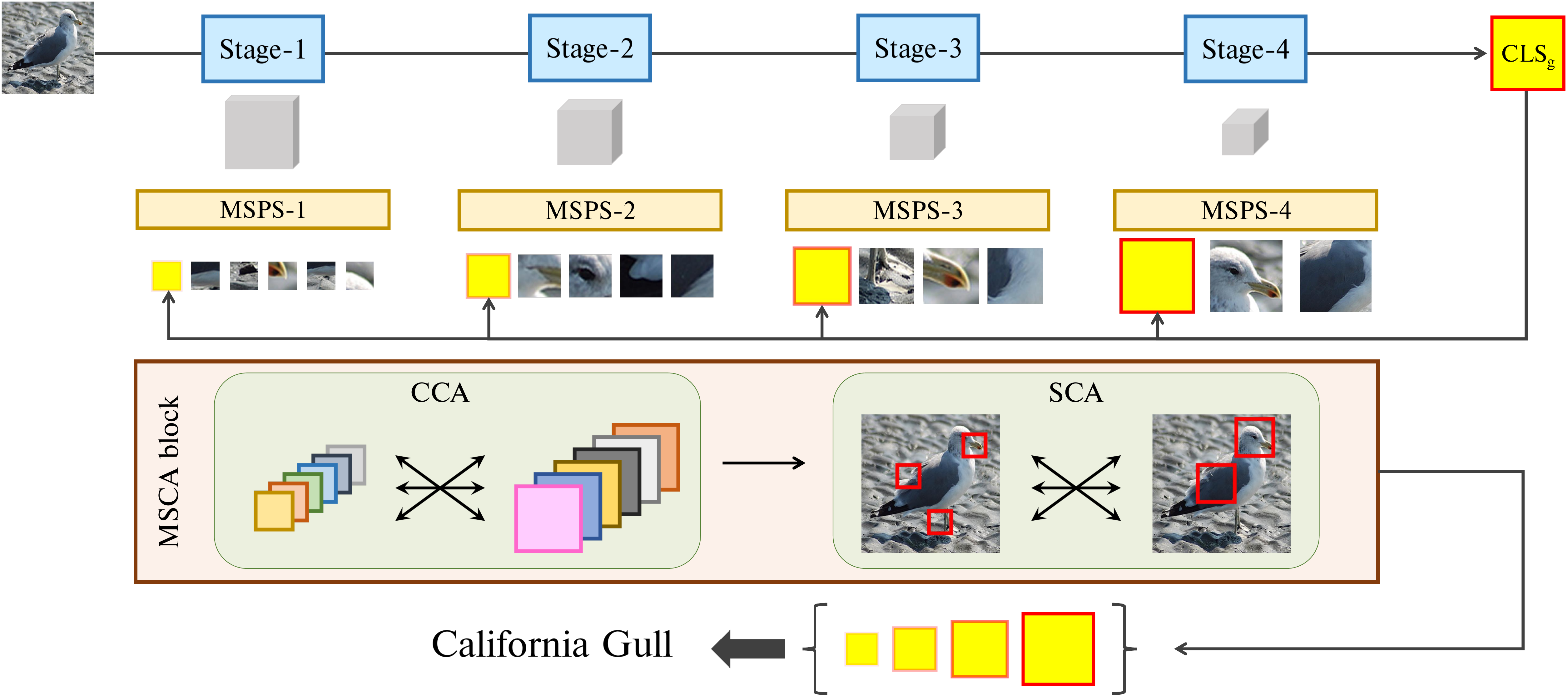}

   \caption{The framework of our M2Former. MSPS conduct patch selection at each stage of the MViT backbone. For each intermediate feature map, several salient patches are selected based on score maps computed using mean activation. At the same time, the global \texttt{CLS} token is transferred to each stage, and the transferred \texttt{CLS} tokens are concatenated with the selected patch sequence. And then, patch sequences are passed through MSCA blocks consisting of CCA and SCA. Finally, the \texttt{CLS} tokens are detached from patch sequence of each stage, and the final prediction is conducted by aggregating them.}
   \label{fig2}
\end{figure*}
Multi-scale features are important for most vision tasks, \textit{e.g.}, object detection~\cite{ETC4, ETC5, MULT4, MULT5, MULT6}, semantic segmentation~\cite{MULT7, MULT8, MULT9}, edge detection~\cite{MULT10, MULT11}, and image classification~\cite{MULT1, MULT2, MULT3}, because visual patterns occur at multi-scales in natural scenes. CNNs naturally learn coarse-to-fine multi-scale features through a stack of convolutional operators~\cite{MULT12, MULT13}, so most research has been proposed to enhance the multi-scale capabilities of CNNs~\cite{MULT3, ETC4, ETC5, MULT14, MULT15}. FPN~\cite{ETC5} introduces a feature pyramid to extract features with different scales from a single image, and fuses them in a top-down way to generate a semantically strong multi-scale feature representation. SPPNet~\cite{MULT3} proposes spatial pyramid pooling to improve backbone networks to model multi-scale features of arbitrary size. Faster R-CNN~\cite{MULT4} proposes region proposal networks that generate object bounding boxes of different scales. FCN~\cite{MULT14} proposes a fully convolutional architecture that generates dense prediction maps for semantic segmentation from hierarchical representations of CNNs.

On the other hand, ViT suffers from being unsuitable to handle multi-scale features due to fixed scale image patches. After the initial patch embedding layer, ViT maintains image patches of the same size, and these single-scale feature maps are not suitable for many vision tasks, especially those requiring pixel-level dense prediction. To alleviate this issue, model architectures that adapt multi-scale feature hierarchies to ViT have recently been proposed. CvT~\cite{TRMS1} adapts convolutional operations to patch embedding layers and attention projection layers to enable feature hierarchies with improved efficiency. MViT~\cite{TRMS2} introduces attention pooling that controls feature resolution by adjusting the pooling stride of queries to implement multi-scale feature hierarchies. PVT~\cite{TRMS4} proposes a progressive shrinking strategy that conducts patch embedding with different patch sizes at the beginning of each stage. SwinT~\cite{TRMS5} generates multi-resolution feature maps by merging adjacent local patches using a patch merging layer. We also focus on feature hierarchies of fine-grained objects using multi-scale vision Transformers. However, we propose additional methods to further boost the multi-scale capability, \textit{i.e.}, MSPS, CTT, and MSCA.
%
%
\section{Our Method}
\label{sec:method}
The overall framework of our method is presented in Fig~\ref{fig2}. First, we use multi-scale vision Transformers (MS-ViT) as our backbone network (Section~\ref{subsec:hvit}). 
After that, multi-scale patch selection (MSPS) is equipped on different stages of MS-ViT to extract multi-scale salient patches (Section~\ref{subsec:msps}). Class token transfer (CTT) aggregates multi-scale patch information by transferring the global \texttt{CLS} token to each stage. Multi-scale cross-attention (MSCA) blocks are used to model spatial-/channel-wise interactions of selected multi-scale patches (Section~\ref{subsec:msca}). Finally, we use additional training strategies for better optimization (Section~\ref{subsec:train}). More details are described as follows.
\subsection{Multi-Scale Vision Transformer}
\label{subsec:hvit}
To enhance the multi-scale capability, we use MS-ViT as our backbone network, specifically the recent Multiscale Vision Transformer (MViT)~\cite{TRMS2, TRMS3}. MViT constructs four-stage pyramid structure for low-level to high-level visual modeling instead of single-scale processing. To produce a hierarchical representation, MViT introduces Pooling Attention (PA), which pools query tensors to control the downsampling factor. We refer the interested reader to the original work~\cite{TRMS2, TRMS3} for details.

Let $X_{0}\in{\mathbb{R}^{h_{0}\times{w_{0}}\times{c_{0}}}}$ denote the input image where $h_{0}$, $w_{0}$, and $c_{0}$ refer to the height, width, and the number of channels, respectively. $X_0$ first goes through a patch embedding layer to produce initial feature maps with a patch size of $4\times{4}$. As the stage deepens, the resolution of the feature maps decreases and the channel dimension increases proportionally. As a result, at each stage $i\in{\{1, 2, 3, 4\}}$, we can extract the feature maps $X_i\in\{X_1, X_2, X_3, X_4\}$ with resolutions ${h_i}\times{w_i}\in\{{\frac{h_0}{4}\times\frac{w_0}{4}}, {\frac{h_0}{8}\times\frac{w_0}{8}}, {\frac{h_0}{16}\times\frac{w_0}{16}}, {\frac{h_0}{32}\times\frac{w_0}{32}}\}$ and channel dimensions $c_i\in\{96, 192, 384, 768\}$. We can also flatten $X_i$ into 1D patch sequence as $X_i\in{\mathbb{R}^{l_i\times{c_{i}}}}$, where $l_i=h_i\times{w_i}$. In fact, after patch embedding, we attach a trainable class token (\texttt{CLS} token) to patch sequence, and all patches $\widetilde{X}_i\in{\mathbb{R}^{\widetilde{l}_i\times{c_{i}}}}$ are fed into consecutive encoder blocks, where $\widetilde{l}_i=l_i+1$. After the last block, the \texttt{CLS} token is detached from the patch sequence and used for class prediction through a linear classifier.
\subsection{Multi-Scale Patch Selection}
\label{subsec:msps}
Single-scale patch selection (SSPS) has limited representations due to its fixed receptive field. Therefore, We propose multi-scale patch selection (MSPS) that extends SSPS to multi-scale. With multiple receptive fields, our proposed MSPS encourages rich representations of objects from deep semantic information to fine-grained details when compared to SSPS. We design MSPS based on the MViT backbone. Specifically, we select salient patches from the intermediate feature maps produced at each stage of MViT. 

Given patch sequence $\widetilde{X}_{i}\in{\mathbb{R}^{\widetilde{l}_i\times{c_{i}}}}$, we start by detaching the \texttt{CLS} token and reshaping it into 2D feature maps to $X_{i}\in{\mathbb{R}^{h_{i}\times{w_{i}}\times{c_{i}}}}$. And then, we group $r\times{r}$ neighboring patches, reshaping  $X_{i}$ into $\Hat{X}_{i}\in{\mathbb{R}^{r^{2}\times\frac{{h_{i}w_{i}}}{r^{2}}\times{c_{i}}}}$. This means $\frac{{h_{i}w_{i}}}{r^{2}}$ neighboring patch groups are generated. Afterwards, we apply a per-group average to merge patch groups, producing $\Hat{X}_{i}\in{\mathbb{R}^{\Hat{l}_i\times{c_{i}}}}$, where $\Hat{l}_i=\frac{{h_{i}w_{i}}}{r^{2}}$. We set $r=2$ to merge patches within a $2\times2$ local region. This merging process removes the redundancies of neighboring patches, which forces MSPS to search for salient patches in wider areas of the image.

Now, we produce a score map $S_i\in{\mathbb{R}^{\Hat{l}_i}}$ using a pre-defined scoring function $\mathcal{S}\left(\cdot\right)$. Then, patches with top-$k$ scores are selected from $\Hat{X}_{i}$, 
\begin{equation}
P_i = \text{MSPS}\left(\Hat{X}_{i};S_i, k_i\right),
\label{equation1}
\end{equation}

\noindent where $P_i\in{\mathbb{R}^{{k_{i}}\times{c_{i}}}}$. We set $k$ differently for each stage to consider hierarchical representations. Since the high-resolution feature maps of the lower stage capture the detailed shape of the object with a small patch size, we set $k$ to be large so that enough patches are selected to sufficiently represent the details of the object. On the other hand, low-resolution feature maps of the higher stage capture the semantic information of objects with a large patch size, so small $k$ is sufficient to represent the overall semantics.

For patch selection, we have to decide how to define the scoring function $\mathcal{S}$. Attention roll-out~\cite{ETC3} has been mainly used as a scoring function for SSPS~\cite{FGTR1,FGTR2}. Attention roll-out aggregates the attention weights of the Transformer blocks through successive matrix multiplications, and the patch selection module selects the most salient patches based on the aggregated attention weights. However, since we use MS-ViT as the backbone, we cannot use attention roll-out because the size of attention weights is different for each stage, even each block. Instead, we propose a simple scoring function based on mean activation, where the score for the $j$-th patch of $\Hat{X}_{i}$ is calculated by:
\begin{equation}
S_i^{j} = \mathcal{S}(\Hat{X}_{i}) = \frac{1}{c_i}\sum_{c=1}^{c_i}\Hat{X}_{i}^{j}\left(c\right),
\label{equation2}
\end{equation}

\noindent where $c$ is the channel index $c\in\{1,2,\dots,c_i\}$. Mean activation measures how strongly the channels in each patch are activated on average. After computing the score map, our MSPS conducts patch selection based on it. This is implemented through top-$k$ and gather operations. We extract $k_i$ patch indices with the highest scores from the $S_i$ through the top-$k$ operation, and patches corresponding to the patch indices $I_i$ are selected from $\Hat{X}_{i}$,
\begin{equation}
\begin{aligned}
&I_i = \text{topkIndex}(S_i;k_i),\\
&P_i = \text{gather}(\Hat{X}_{i};I_i),
\end{aligned}
\label{equation3}
\end{equation}

\noindent where $I_i\in{\mathbb{N}^{{k_{i}}}}$, and $P_i\in{\mathbb{R}^{{k_{i}}\times{c_i}}}$.

\subsection{Class Token Transfer}
\label{subsec:ctt}
Through MSPS, we can extract salient patches from each stage, $\mathbf{P}=\{P_1, P_2, P_3, P_4\}$. In Section~\ref{subsec:msps}, the \texttt{CLS} token is detached from the patch sequence before MSPS at each stage. The simplest way to reflect the selected multi-scale patches in the model decisions is to concatenate the detached \texttt{CLS} token $\texttt{CLS}_i$ with the $P_{i}$ again and feed it into a few additional ViT blocks, consisting of multi-head self-attention (MSA) and feed-forward networks (FFN):
\begin{equation}
\begin{aligned}
&\widetilde{P}_i = \text{concat}(P_{i}, \texttt{CLS}_i),\\
&\widetilde{O}_i = \text{FFN}(\text{MSA}(\widetilde{P}_i)),
\end{aligned}
\label{equation4}
\end{equation}

\noindent where $\widetilde{P}_i$, $\widetilde{O}_i\in{\mathbb{R}^{\widetilde{k}_{i}\times{c_{i}}}}$, and $\widetilde{k}_i=k_i+1$. Finally, predictions for each stage are computed by extracting the $\texttt{CLS}_i$ from $\widetilde{O}_i$ and connecting the linear classifier. It should be noted that the $\texttt{CLS}_i$ is shared by all stages: the set of $\texttt{CLS}_i$ is derived from the global \texttt{CLS} token and it is detached with different dimensions at each stage. This means that the stage-specific multi-scale information is shared to some extent through $\texttt{CLS}_i$. However, the current sharing method may cause inconsistency between stage features because the detached $\texttt{CLS}_i$ does not equally utilize the representational power of the network. For example, $\texttt{CLS}_1$ is detached right after stage-1 and it will always lag behind $\texttt{CLS}_4$, which shares the same root as $\texttt{CLS}_1$ but utilizes representations of all stages.

To this end, we introduce a class token transfer (CTT) strategy that aggregates multi-scale information more effectively. The core idea is to use the \texttt{CLS} token transferred from the global \texttt{CLS} token $\texttt{CLS}_g$ rather than using the detached $\texttt{CLS}_i$ at each stage. It should be noted that $\texttt{CLS}_g$ is equal to $\texttt{CLS}_4$, so $\texttt{CLS}_{g}\in{\mathbb{R}^{c_{4}}}$. We transfer the $\texttt{CLS}_g$ according to the dimension of each stage through a projection layer consisting of two linear layers along with Batch Normalization (BN) and ReLU activation:
\begin{equation}
\overline{\texttt{CLS}}_i = W_i^1\left(\text{ReLU}\left(\text{BN}\left(W_i^0\texttt{CLS}_g\right)\right)\right),
\label{equation5}
\end{equation}

\noindent where $W_i^0\in{\mathbb{R}^{2c_{i}\times{c_{4}}}}$, $W_i^1\in{\mathbb{R}^{c_{i}\times{2c_i}}}$ are the weight matrices, and $\overline{\texttt{CLS}}_i$ is the transferred $\texttt{CLS}_g$ in stage $i\in\{1, 2, 3\}$. Now (\ref{equation4}) is reformulated as:
\begin{equation}
\begin{aligned}
&\widetilde{P}_i = \text{concat}(P_{i}, \overline{\texttt{CLS}}_i),\\
&\widetilde{O}_i = \text{FFN}(\text{MSA}(\widetilde{P}_i)).
\end{aligned}
\label{equation6}
\end{equation}

\noindent Compared to conventional approaches, CTT guarantees consistency between stage features as it uses \texttt{CLS} tokens utilized with the same representational power. Each stage encodes stage-specific patch information into a globally updated \texttt{CLS} token. CTT is similar to the top-down pathway~\cite{ETC5, MULT5}: it combines high-level representations of objects with multi-scale representations of lower layers to generate richer network-level representations.
\subsection{Multi-Scale Cross-Attention}
\label{subsec:msca}
\begin{figure}[t]
  \centering
   \includegraphics[width=1.0\linewidth]{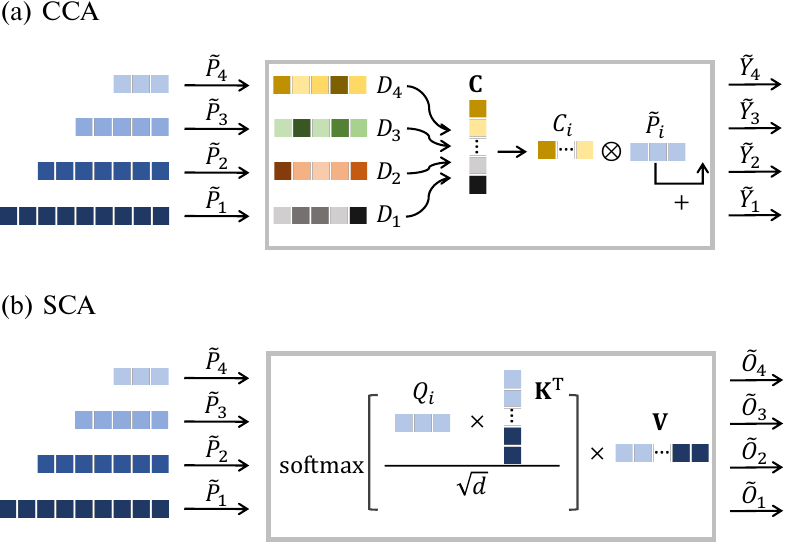}

   \caption{CCA and SCA constituting the MSCA block. (a) CCA recalibrates the channels of each stage-specific patches based on their cross-scale channel interdependencies. (b) For the same purpose, SCA captures spatial-wise interdependencies of selected multi-scale patches.}
   \label{fig3}
\end{figure}
Although CTT can aggregate multi-scale patch information from all stages, it cannot model direct interactions between multi-scale patches, which indicates how interrelated they are. Therefore, we propose multi-scale cross-attention (MSCA) to model the interactions between multi-scale patches. 

MSCA takes $\widetilde{\mathbf{P}}=\{\widetilde{P}_1, \widetilde{P}_2, \widetilde{P}_3, \widetilde{P}_4\}$ as input and models the interactions between selected multi-scale salient patches. Specifically, MSCA consists of channel cross-attention (CCA) and spatial cross-attention (SCA), so (\ref{equation6}) is reformulated as:
\begin{equation}
\widetilde{\mathbf{O}}=\text{MSCA}(\widetilde{\mathbf{P}}) = \text{SCA}(\text{CCA}(\widetilde{\mathbf{P}})),
\label{equation7}
\end{equation}

\noindent where $\widetilde{\mathbf{O}}=\{\widetilde{O}_1, \widetilde{O}_2, \widetilde{O}_3, \widetilde{O}_4\}$.
\subsubsection{Channel Cross-Attention}
\label{subsubsec:cca}
Exploring feature channels has been very important in many vision tasks because feature channels encode visual patterns that are strongly related to foreground objects~\cite{FGWS2, FGWS6, FGET3, FGET15, FGET16}. Many studies have been proposed to enhance the representational power of a network by explicitly modeling the interdependencies between the feature channels~\cite{ETC11, ETC12, ETC13, ETC14, ETC15}. In the same vein, we propose CCA to further enhance the representational richness of multi-scale patches by explicitly modeling their cross-scale channel interactions. 

We illustrate CCA in Fig.~\ref{fig3} (a). First, we apply global average pooling (GAP) to $\widetilde{P}_i$ to obtain a global channel descriptor $D_i\in{\mathbb{R}^{c_i}}$ for each stage. The $c$-th element of $D_i$ is calculated by:
\begin{equation}
D_{i}^{c}=\frac{1}{\widetilde{k}_{i}}\sum_{j=1}^{\widetilde{k}_{i}}{\widetilde{P}_{i}^{c}\left(j\right)},
\label{equation8}
\end{equation}

\noindent where $j$ is the patch index $j\in\{1,2,\dots,\widetilde{k}_{i}\}$. From the stage-specific channel descriptors, we compute the channel attention score as follows:
\begin{equation}
\begin{aligned}
&\mathbf{D}=\text{concat}\left(D_1, D_2, D_3, D_4\right)\in{\mathbb{R}^{c}},\\
&\mathbf{C}=\text{sigmoid}\left(W^{c,1}\text{ReLU}\left(\text{BN}\left(W^{c,0}\mathbf{D}\right)\right)\right)\in{\mathbb{R}^{c}},
\end{aligned}
\label{equation9}
\end{equation}

\noindent where $c=\sum{c_i}$, $W^{c,0}\in{\mathbb{R}^{{\frac{c}{2}}\times{c}}}$, and $W^{c,1}\in{\mathbb{R}^{c\times\frac{c}{2}}}$. We then split $\mathbf{C}$ back into $C_i\in{\mathbb{R}^{c_i}}$ and recalibrate the channels of $\widetilde{P}_i$ as follows:
\begin{equation}
\widetilde{Y}_{i}=\widetilde{P}_i\otimes{C_i}+\widetilde{P}_i,
\label{equation10}
\end{equation}

\noindent where $\otimes$ indicates element-wise multiplication. In (\ref{equation9}), we compute the channel attention score by aggregating the channel descriptors of all multi-scale patches. It captures channel-dependencies in a cross-scale way and reflects them back to each stage-specific channel information.
\subsubsection{Spatial Cross-Attention}
\label{subsubsec:sca}
In addition to channel-wise interactions, we can compute the spatial-wise interdependencies of selected multi-scale patches. To this end, we propose SCA, which is a multi-scale extension of MSA~\cite{ETC1, ETC2}.

We illustrate SCA in Fig.~\ref{fig3} (b). First, We compute query, key, value tensors $Q_i$, $K_{i}$, $V_{i}$ for every $\widetilde{Y}_i$,
\begin{equation}
Q_i=\widetilde{Y}_{i}W_i^{Q},{\quad}K_{i}=\widetilde{Y}_{i}W_i^{K},{\quad}V_{i}=\widetilde{Y}_{i}W_i^{V},
\label{equation11}
\end{equation}

\noindent where $W_i^{Q}$, $W_i^{K}$, $W_i^{V}\in{\mathbb{R}^{c_{i}\times{d}}}$, and $Q_i$, $K_i$, $V_i\in{\mathbb{R}^{\widetilde{k}_{i}\times{d}}}$. After that, we concatenate the $K_i$ and $V_i$ of all stages to generate global key and value tensors $\mathbf{K}$, $\mathbf{V}$,
\begin{equation}
\begin{aligned}
&\mathbf{K}=\text{concat}\left(K_{1}, K_{2}, K_{3}, K_{4}\right)\in{\mathbb{R}^{\widetilde{k}\times{d}}},\\
&\mathbf{V}=\text{concat}\left(V_{1}, V_{2}, V_{3}, V_{4}\right)\in{\mathbb{R}^{\widetilde{k}\times{d}}},
\end{aligned}
\label{equation12}
\end{equation}

\noindent where $\widetilde{k}=\sum{\widetilde{k}_i}$. Now, we can compute self-attention for $Q_i$, $\mathbf{K}$, $\mathbf{V}$, and single linear layer is used to restore the dimension,
\begin{equation}
\begin{aligned}
&\widetilde{A}_i=\text{Softmax}\left(Q_{i}\mathbf{K}^{T}/\sqrt{d}\right)\mathbf{V},\\
&\widetilde{O}_i=\widetilde{A}_iW_i^{s},
\end{aligned}
\label{equation13}
\end{equation}

\noindent where $\widetilde{A}_{i}\in{\mathbb{R}^{\widetilde{k}_{i}\times{d}}}$, $W_i^{s}\in{\mathbb{R}^{d\times{c_i}}}$, and $\widetilde{O}_{i}\in{\mathbb{R}^{\widetilde{k}_{i}\times{c_i}}}$. SCA is also implemented in a multi-head manner~\cite{ETC2}. For global key and value, SCA captures how strongly multi-scale patches interact spatially with each other. Specifically, SCA models how large-scale semantic patches decompose into more fine-grained views, and conversely, how small-scale fine-grained patches can be identified in more global views.
\subsection{Training}
\label{subsec:train}
After the MSCA block, we can extract $\overline{\texttt{CLS}}_i$ from $\widetilde{O}_{i}$ and compute the class prediction $y_i$ using a linear classifier. In addition, we can compute $y_{con}$ by concatenating all $\overline{\texttt{CLS}}_i$ tokens. For model training, we compare every $\mathbf{y}=\{y_1, y_2, y_3, y_4, y_{con}\}$ for the ground-truth label $\hat{y}$,
\begin{equation}
\mathcal{L}=\sum_{y \in\mathbf{y}}{\sum_{t=1}^{n}{-\hat{y}^{t}\log{y^{{t}}}}},
\label{equation14}
\end{equation}

\noindent where $n$ is the total number of classes, and $t$ denotes the element index of the label. To improve model generalization and encourage diversity of representations from specific stages, we employ soft supervision using label smoothing~\cite{ETC6, FGWS1}. We modify the one-hot vector $\hat{y}$ as follows:
\begin{equation}
\hat{y}_{\alpha}[t]=
    \begin{cases}
      \alpha & t=\hat{t} \\
      \frac{1-\alpha}{n} & t\neq\hat{t}
    \end{cases},
\label{equation15}
\end{equation}

\noindent where $\hat{t}$ denotes index of the ground-truth class, and $\alpha$ denotes a smoothing factor $\alpha\in[0,1]$. $\alpha$ controls the magnitude of the ground-truth class. As a result, the different predictions are supervised with different labels during training. We set $\alpha$ to increase in equal intervals by $0.1$ from $0.6$ to $1$, so $y_1$ has the smallest $\alpha=0.6$.

For inference, we conduct a final prediction considering all of $\mathbf{y}$,
\begin{equation}
y_{all}=\sum_{y \in\mathbf{y}}{y},
\label{equation16}
\end{equation}
\noindent where the maximum entry in the $y_{all}$ corresponds to the class prediction.

%
%
\section{Experiments}
\label{sec:experiment}
\begin{table}[t]
\centering
\caption{Comparison with the state-of-the-art methods on CUB.}
\begin{tabular}{l|c|c}
\hline
\multicolumn{1}{c|}{method} & backbone    & accuracy (\%) \\ \hline\hline
RA-CNN~\cite{FGWS13}        & VGG19       & 85.3          \\
NTS-Net~\cite{FGET7}        & ResNet50    & 87.5          \\
Cross-X~\cite{FGET8}        & ResNet50    & 87.7          \\
DBTNet~\cite{FGET5}         & ResNet101   & 88.1          \\
DF-GMM~\cite{FGWS5}         & ResNet50    & 88.8          \\
PMG~\cite{FGET1}            & ResNet50    & 89.6          \\
API-Net~\cite{FGET4}        & DenseNet161 & 90.0          \\
P2P-Net~\cite{FGWS1}        & ResNet50    & 90.2          \\ \hline
TransFG~\cite{FGTR1}        & ViT-B       & 91.7          \\ 
RAMS-Trans~\cite{FGTR3}     & ViT-B       & 91.3          \\
FFVT~\cite{FGTR4}           & ViT-B       & 91.6          \\
DCAL~\cite{FGTR2}           & R50-ViT-B   & 92.0          \\ \hline
M2Former (ours)              & MViTv2-B    & \textbf{92.4} \\ \hline
\end{tabular}
\label{table1}
\end{table}
\begin{table}[t]
\centering
\caption{Comparison with the state-of-the-art methods on NABirds.}
\begin{tabular}{l|c|c}
\hline
\multicolumn{1}{c|}{method} & backbone    & accuracy (\%) \\ \hline\hline
MaxEnt~\cite{FGET10}        & DenseNet161 & 83.0          \\
Cross-X~\cite{FGET8}        & ResNet50    & 86.4          \\
PAIRS~\cite{FGET11}         & ResNet50    & 87.9          \\
API-Net~\cite{FGET4}        & DenseNet161 & 88.1          \\
PMGv2~\cite{FGET9}          & ResNet101   & 88.4          \\
CS-Parts~\cite{FGET12}      & ResNet50    & 88.5          \\
MGE-CNN~\cite{FGET13}       & ResNet101   & 88.6          \\
FixSENet-154~\cite{FGET14}  & SENet154    & 89.2          \\ \hline
ViT~\cite{ETC1}             & ViT-B       & 89.9          \\
TransFG~\cite{FGTR1}        & ViT-B       & 90.8          \\ \hline
M2Former (ours)              & MViTv2-B    & \textbf{91.1} \\ \hline
\end{tabular}
\label{table2}
\end{table}
In this section, we evaluate our proposed M2Former on widely used fine-grained benchmarks. In addition, we conduct ablation studies and further analyses to validate the effectiveness of our method. More details are described as follows.
\subsection{Datasets}
\label{subsec:dataset}
We use two FGVR datasets to evaluate the proposed method: Caltech-UCSD Birds (CUB)~\cite{DATA1}, and NABirds~\cite{DATA7}. The \textbf{CUB} dataset consists of 11,788 images and 200 bird species. All images are split into 5,994 for training and 5,794 for testing. The \textbf{NABirds} is a larger dataset, consisting of 48,562 images and 555 classes. All images are split into 23,929 for training and 24,633 for testing. Most of our experiments are conducted on CUB.
\subsection{Implementation Details}
\label{subsec:detail}
We use MViTv2-B~\cite{TRMS3} pre-trained on ImageNet21K~\cite{ETC7} as our backbone network. We add MSPS to every stage of MViTv2-B. After patch selection, selected patches pass through one MSCA block. We empirically set $\mathbf{k}=\{k_1, k_2, k_3, k_4\}$, the number of patches selected for each stage, to $\{6, 18, 54, 162\}$. Our training recipe follows the same as recent work~\cite{FGTR1}. The model is trained for a total of 10,000 iterations, and an SGD optimizer with momentum of 0.9 and weight decay of 0 is used. The batch size is set to 16, and the initial learning rate is set to 0.03. The learning rate has a cosine decay schedule~\cite{ETC9}. For augmentations, raw images are first resized into $600\times600$ followed by cropping into $448\times448$. We use random cropping for training, and center cropping for testing. Random horizontal flipping is adapted to training images. We implement the whole model with the PyTorch framework on three NVIDIA A5000 GPUs.
\subsection{Main Results}
\label{subsec:main}
We compare our M2Former with state-of-the-art FGVR methods including ViT-based and CNN-based models on each dataset.
\subsubsection{Results on CUB}
\label{subsubsec:cub}
First, the evaluation results on CUB are presented in Table~\ref{table1}. As shown in Table~\ref{table1}, our M2Former obtained a top-1 accuracy of 92.4\%, which significantly outperforms CNN-based methods. Especially, M2Former improves recent P2P-Net~\cite{FGWS1} by 2.2\% higher. In addition, M2Former achieved higher accuracy compared to ViT-based models using SSPS. Specifically, M2Former outperforms TransFG~\cite{FGTR1}, RAMS-Trans~\cite{FGTR3}, FFVT~\cite{FGTR4}, and DCAL~\cite{FGTR2} by 0.7\%, 1.1\%, 0.8\%, and 0.4\%  higher respectively. These results indicate that our proposed MSPS encourages enhanced representations compared to SSPS.
\subsubsection{Results on NABirds}
\label{subsubsec:nabirds}
The evaluation results on NABirds are presented in Table~\ref{table2}. Our proposed M2Former obtains the top-1 accuracy of 91.1\% on NABirds. Compared to CNN-based models, our M2Former shows significantly improved performance. For example, M2Former improves PMGv2~\cite{FGET9} by 2.7\% higher. In addition, our method outperforms other ViT-based models. For example, M2Former improves ViT~\cite{ETC1} by 1.2\% higher and TransFG~\cite{FGTR1} by 0.3\% higher.
\subsection{Ablation Study}
\label{subsec:ablation}
We analyzed each component of the proposed M2Former through the ablation study. All experiments are conducted on CUB dataset.
\begin{table}[t]
\centering
\caption{Effect of the proposed modules on CUB.}
\begin{tabular}{c|cccc|c}
\hline
index & MSPS          & CTT          & CCA          & SCA          & accuracy (\%) \\ \hline\hline
(a)   &               &              &              &              & 91.6          \\
(b)   & \checkmark    &              &              &              & 91.6          \\
(c)   & \checkmark    & \checkmark   &              &              & 92.1          \\
(d)   & \checkmark    & \checkmark   & \checkmark   &              & 92.2          \\
(e)   & \checkmark    & \checkmark   &              & \checkmark   & 92.1          \\
(f)   & \checkmark    & \checkmark   & \checkmark   & \checkmark   & \textbf{92.4} \\ \hline
\end{tabular}
\label{table3}
\end{table}
\begin{table}[t]
\centering
\caption{Effect of the number of MSCA blocks on CUB.}
\begin{tabular}{c|c}
\hline
num. blocks & accuracy (\%) \\ \hline\hline
1           & \textbf{92.4} \\
2           & 92.1          \\
3           & 92.2          \\
4           & 92.2          \\ \hline
\end{tabular}
\label{table4}
\end{table}
\begin{table}[t]
\centering
\caption{Effect of the number of selected patches on CUB.}
\begin{tabular}{c|c}
\hline
$\mathbf{k}$        & accuracy (\%) \\ \hline\hline
\{6, 8, 10, 12\}    & 92.2          \\
\{6, 12, 24, 48\}   & 92.2          \\
\{6, 18, 54, 162\}  & \textbf{92.4} \\
\{7, 28, 112, 448\} & 92.1          \\ \hline
\end{tabular}
\label{table5}
\end{table}
\begin{table}[t]
\centering
\caption{Comparison of different backbone networks on CUB.}
\begin{tabular}{l|c}
\hline
\multicolumn{1}{c|}{method} & accuracy (\%) \\ \hline\hline
CvT-21~\cite{TRMS1}         & 89.3          \\
CvT-21$_{\text{MSPS}}$      & \textbf{90.0} \\ \hline
SwinT-B~\cite{TRMS5}        & 90.6          \\
SwinT-B$_{\text{MSPS}}$     & \textbf{91.3} \\ \hline
\end{tabular}
\label{table6}
\end{table}
\subsubsection{Ablation on Proposed Modules}
\label{subsubsec:modules}
We investigate the influence of the proposed modules (\textit{i.e.}, MSPS, CTT, CCA, SCA) included in the M2Former architecture. The results are presented in Table~\ref{table3}. The pure MViTv2-B baseline obtained top-1 accuracy of 91.6\% on CUB (Table~\ref{table3} (a)). When MSPS was added, we did not find any noticeable improvement (Table~\ref{table3} (b)). This indicates that effectively incorporating selected patches into model decisions is more important than patch selection itself. For this purpose, CTT allows multi-scale patch information to be shared across the entire network through transferred global \texttt{CLS} tokens. Indeed, with CTT, we improve the baseline by 0.5\% higher (Table~\ref{table3} (c)). In addition, CCA and SCA capture more direct interactions between multi-scale patches, but using CCA or SCA alone was not effective for improvement (Table~\ref{table3} (d) and (e)). On the other hand, when using both CCA and SCA (full MSCA block) we obtain the highest top-1 accuracy of 92.4\%, which improves baseline by 0.8\% higher (Table~\ref{table3} (f)). This means that spatial-/channel-level interactions of multi-scale patches are strongly correlated and need to be considered simultaneously.
\subsubsection{Number of MSCA Blocks}
\label{subsubsec:blocks}
Table~\ref{table4} shows the results of the ablation experiment on the number of our MSCA blocks. The results show that using a single MSCA block performs best and increasing the number of MSCA blocks no longer yields a meaningful improvement. This means that a single MSCA block is sufficient to model the interactions of multi-scale patches. Moreover, using one MSCA block is efficient as it only introduces small extra parameters.
\subsubsection{Number of Selected Patches}
\label{subsubsec:patches}
\begin{table}[t]
\centering
\caption{Comparison of different CTT methods on CUB.}
\begin{tabular}{c|c}
\hline
method        & accuracy (\%) \\ \hline\hline
global pool   & 91.5          \\
simple attach & 91.9          \\
CTT w/ 1-MLP  & 92.1          \\
CTT w/ 2-MLP  & \textbf{92.4} \\ \hline
\end{tabular}
\label{table7}
\end{table}
Table~\ref{table5} shows the influences from the number of selected patches. We define it as $\mathbf{k}=\{k_1, k_2, k_3, k_4\}$, where $k_i$ is the number of patches selected through patch selection at stage $i$. $k_i$ is set differently for each stage. As shown in Table~\ref{table5}, the larger the overall number of selected patches, the better the performance in general. However, after it grows to some extent, increasing selected patches is insignificant and caused unnecessary computations. We empirically found the optimal $\mathbf{k}=\{6,8,54,162\}$. 
\subsubsection{Different Backbone Networks}
\label{subsubsec:backbone}
In Table~\ref{table6}, we analyze whether our method provides consistent effects on different backbone networks. We use CvT-21~\cite{TRMS1} and SwinT-B~\cite{TRMS5} as backbone architectures and compare them to their baseline. Both are initialized with pre-trained weights from ImageNet21K~\cite{ETC7}. The proposed modules are added to all three stages for CvT-21 and all four stages for SwinT-B. We follow the training recipe from recent work~\cite{FGTR1}. It should be noted that CTT could not be applied to the SwinT-B backbone as it does not use a \texttt{CLS} token. In addition, SwinT-B uses an input resolution of $384\times384$: raw images are first resized into $510\times510$ followed by cropping into $384\times384$. As shown in Table~\ref{table6}, Cvt-21 and SwinT-B baselines obtained top-1 accuracies of 89.3\% and 90.6\%, respectively. When we added the proposed modules, we could obtain top-1 accuracies of 90.0\% and 91.3\%, which are both 0.7\% higher than their pure counterparts. This suggests that selecting multi-scale salient patches and modeling their interactions is important for fine-grained recognition, and our method is beneficial for that purpose.
\subsection{Further Analysis}
\label{subsec:fa}
\begin{figure}[t]
  \centering
   \includegraphics[width=1.0\linewidth]{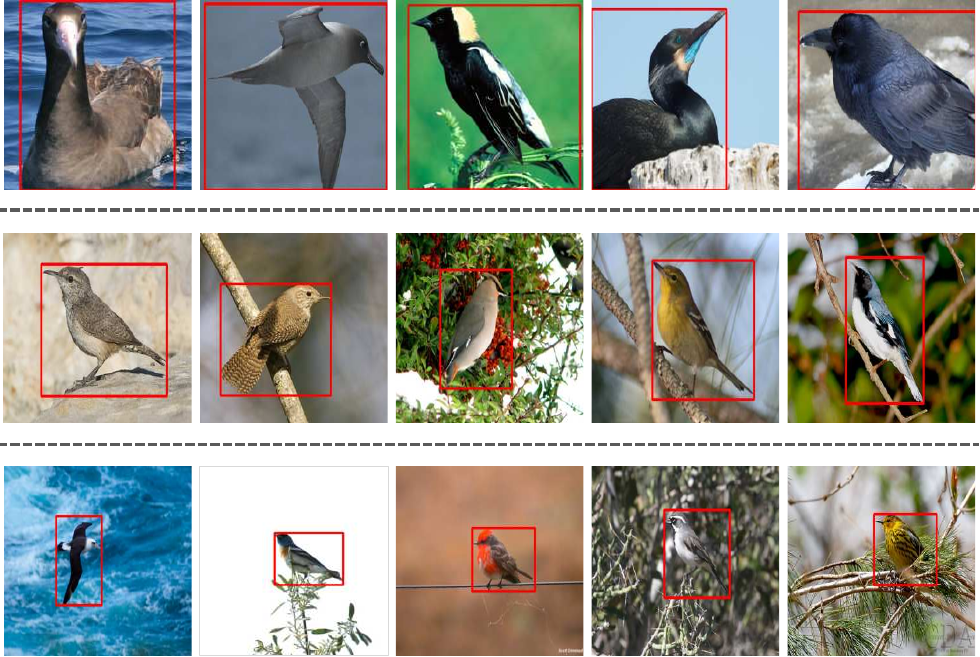}

   \caption{Example images for different scale objects. Following COCO~\cite{ETC10}, we classify objects into three scales: large, medium, and small, according to their bounding box size. The first row shows example images belonging to the large category (ob$_{l}$). The second row shows example images belonging to the medium category (ob$_{m}$). The third row shows example images belonging to the small category (ob$_{s}$).}
   \label{fig4}
\end{figure}
\begin{table*}[t]
\centering
\caption{Comparison of different MSPS stages on CUB.}
\begin{tabular}{l|cccc|lll|l}
\hline
\multicolumn{1}{c|}{\multirow{2}{*}{method}} & \multicolumn{4}{c|}{stages for patch selection}  & \multicolumn{4}{c}{accuracy (\%)}                                                                       \\ \cline{2-9} 
\multicolumn{1}{c|}{}    & stage1        & stage2          & stage3        & stage4 & \multicolumn{1}{c}{ob$_{l}$} & \multicolumn{1}{c}{ob$_{m}$}  & \multicolumn{1}{c}{ob$_{s}$} & \multicolumn{1}{|c}{total} \\ \hline\hline
TransFG~\cite{FGTR1}     & \multicolumn{4}{c|}{\textit{SSPS at last block}}                   &                  
91.7                              & 91.7                             & 90.1                             &        
91.3 \\\hline
M2Former$_{\text{none}}$ &               &                 &               &                  & 92.3\scriptsize{(+0.6)}           & 91.8\scriptsize{(+0.1)}          & 90.5\scriptsize{(+0.4)}          & 91.6\scriptsize{(+0.3)} \\ 
M2Former$_{4}$           &               &                 &               & \checkmark       & 91.9\scriptsize{(+0.2)}           & 91.7\scriptsize{(+0.0)}          & 90.6\scriptsize{(+0.5)}          & 91.5\scriptsize{(+0.2)} \\
M2Former$_{3,4}$         &               &                & \checkmark     & \checkmark       & 92.9\scriptsize{(+1.2)}           & 91.9\scriptsize{(+0.2)}          & 90.8\scriptsize{(+0.7)}          & 91.9\scriptsize{(+0.6)} \\
M2Former$_{2,3,4}$       &               & \checkmark     & \checkmark     & \checkmark       & 92.9\scriptsize{(+1.2)}           & 92.4\scriptsize{(+0.7)}          & 91.4\scriptsize{(+1.3)}          & 92.3\scriptsize{(+1.0)} \\
M2Former$_{\text{full}}$ & \checkmark    & \checkmark     & \checkmark     & \checkmark       & \textbf{92.9}\scriptsize{(+1.2)}  & \textbf{92.6}\scriptsize{(+0.9)} & \textbf{91.4}\scriptsize{(+1.3)} & \textbf{92.4}\scriptsize{(+1.1)} \\ \hline
\end{tabular}
\label{table8}
\end{table*}
In this section, we conduct additional experiments to further validate the effectiveness of our proposed methods. All experiments are conducted on CUB dataset.
\subsubsection{Contributions from CTT}
\label{subsubsec:ctt}
In Table~\ref{table7}, we examine the contributions from CTT. First, we start by not using the \texttt{CLS} token. In this setup, the patch sequence selected at each stage does not contain any \texttt{CLS} tokens. Instead, selected patches that pass through the MSCA block are projected as feature vectors with global average pooling (GAP), and the features are used for final prediction through a linear layer. This is noted as the `global pool' in Table~\ref{table7}. As a result, the global pool obtains a top-1 accuracy of 91.5\%, which lags behind the MViTv2-B baseline (91.6\% in Table~\ref{table3}). We conjecture that this low accuracy is due to the lack of a shared intermediary (\textit{i.e.}, \texttt{CLS} token) to aggregate the information of the selected multi-scale patches, even when using the MSCA block. 

Then, we now initialize the \texttt{CLS} token of the backbone, and the patch sequence selected in each stage is concatenated with the \texttt{CLS} token of the same dimension that was detached before patch selection (as in Section~\ref{subsec:ctt} (\ref{equation4})). This is simply re-attaching the \texttt{CLS} token that was detached before MSPS, so it is noted `simple attach' in Table~\ref{table7}. As a result, the simple attach obtains an accuracy of 91.9\%, which improve the global pool by 0.4\% higher. We argue that this improvement comes from that simple attach can share information between multi-scale patches through the \texttt{CLS} tokens. Since the global \texttt{CLS} token is detached with different channel dimensions at each stage, stage-specific patch information can be shared through the global \texttt{CLS} token to generate richer representations. 

As discussed in Section~\ref{subsec:ctt}, we propose CTT to enhance multi-scale feature sharing. In this setup, each stage uses a 'transferred' global \texttt{CLS} token rather than a simply 'detached' one. This can be implemented using a single projection layer to match its dimensions (noted `CTT w/ 1-MLP' in Table~\ref{table7}). As a result, it obtains an accuracy of 92.1\%, which improve the simple attach by 0.2\% higher. Compared to using a detached \texttt{CLS} token at an intermediate stage, transferring the globally updated \texttt{CLS} token is more effective in that it can inject local information of each stage to the object's deep semantic information, utilizing the same representational power of the network. Finally, CTT performs best when using 2-layer MLP with non-linearity (noted `CTT w/ 2-MLP' in Table~\ref{table7}).
\subsubsection{Contributions from MSPS}
\label{subsubsec:msps}
\begin{figure*}[t]
  \centering
   \includegraphics[width=1.0\linewidth]{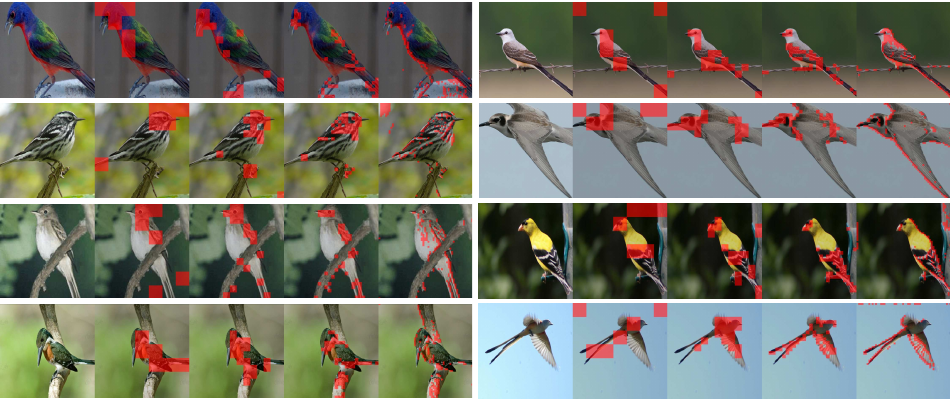}

   \caption{Visualization results of the selected patches when MSPS was conducted at each stage. In each subfigure, the first column shows the original image, and the second to fifth columns show the patches selected from stage-4 to stage-1. The selected patches are marked with red rectangles.}
   \label{fig5}
\end{figure*}
\begin{figure}[t]
  \centering
   \includegraphics[width=1.0\linewidth]{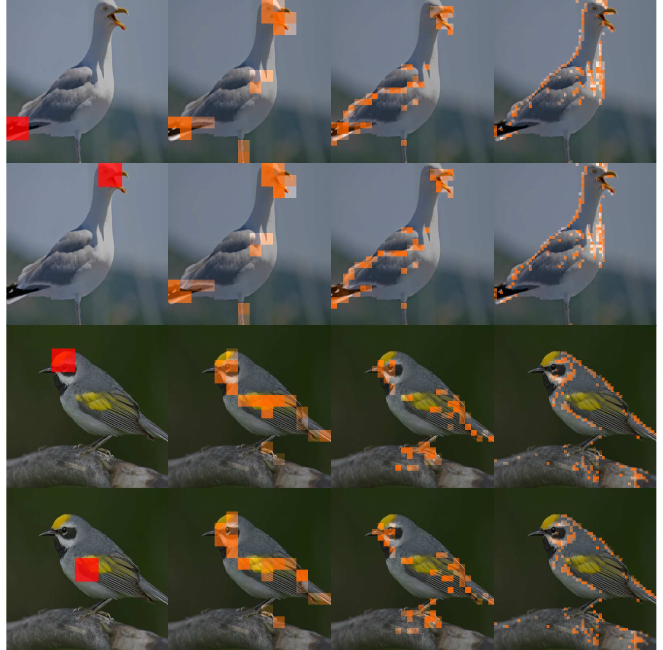}

   \caption{Visualization results of cross-attention maps from SCA. The first column shows the sampled stage-4 patch as a red rectangle. The second to fourth columns show the attention maps between sampled patches and selected patches from stage-3 to stage-1 as orange rectangles. Brighter rectangles mean higher attention scores.}
   \label{fig6}
\end{figure}
In the earlier section, we pointed out that SSPS is suboptimal because it is difficult to deal with scale changes. We now show that our proposed MSPS can consistently improve performance on different object scales. 

First, we need to classify a given set of objects according to their scale. Following COCO~\cite{ETC10}, we categorize all objects into large, medium, and small-sized objects according to their bounding box size. We compute the size of the bounding box using the given box coordinates and sort them in ascending order. Then, we compute the quartiles for sorted bounding box sizes. Finally, we classify an object as a small-sized (ob$_{s}$) object if its bounding box size is less than the first quartile, as a large-sized object (ob$_{l}$) if its bounding box size is greater than the third quartile, and as a medium-sized (ob$_{m}$) object otherwise. Specifically, 25\% of objects are small, 50\% are medium, and 25\% are large. Example images for each category are shown in Fig.~\ref{fig4}.

We train five model variants with different MSPS stages: M2Former$_{\text{none}}$, M2Former$_{4}$, M2Former$_{3,4}$, M2Former$_{2,3,4}$, M2Former$_{full}$. It should be noted that M2Former$_{\text{none}}$ is exactly the same as the MViTv2-B baseline. For comparison, we trained TransFG~\cite{FGTR1}, which conducts SSPS at the last encoder block.

The results are presented in Table~\ref{table8}. First, TransFG obtained a total top-1 accuracy of 91.3\%, which obtained accuracies of 91.7\% for ob$_{l}$, 91.7\% for ob$_{m}$, and 90.1\% for ob$_{s}$. The performance of ob$_{s}$ seems to lag behind ob$_{l}$ and ob$_{m}$. M2Former$_{\text{none}}$ obtains a total top-1 accuracy of 91.6\%, which outperforms TransFG by 0.3\% higher. This means that we can achieve satisfactory results simply by using MS-ViT. Additionally, the improvement was prominent in ob$_{l}$ and ob$_{s}$ (0.6\% and 0.4\% higher, respectively), indicating that multi-scale features are important in mitigating scale variability. 

When we add MSPS to stage-4 (M2Former$_{4}$), we obtain a top-1 accuracy of 91.5\%, which is 0.2\% higher than the baseline. M2Former$_{4}$ is almost identical to SSPS as it selects only single-scale patches in a single stage. However, M2Former$_{4}$ obtained a lower total top-1 accuracy than M2Former$_{\text{none}}$, and it generally results in lower accuracy at all scales. This is also in contrast to previous findings~\cite{FGTR1} where performance was improved when patch selection was conducted at the last block. 

However, when adding the MSPS in stage-3 (M2Former$_{3,4}$), we obtained a top-1 accuracy of 91.9\%, which outperforms the SSPS baseline by 0.6\% higher. The improvements were found at all object scales, but were noticeable at ob$_{l}$ (1.2\% higher compared to the SSPS baseline). This means that salient patches from stage-3 (along with selected patches from stage-4) provide the definitive cue for large objects. 

We can see this improvement even when adding MSPS in stage-2 (M2Former$_{2,3,4}$). Especially, adding MSPS in stage-2 leads to improvements for ob$_{m}$ and ob$_{s}$. Compared to the SSPS baseline, it is 0.7\% higher for ob$_{m}$, 1.3\% higher for ob$_{s}$, and 1.0\% higher for total accuracy. This indicates that the finer-grained object features extracted at stage-2 enhance representations for small/medium-sized objects. 

Finally, when adding MSPS to stage-1 (M2Former$_{full}$), we obtained the highest total top-1 accuracy of 92.4\%, with a slight further improvement in ob$_{m}$. In summary, MSPS from high to low stages models a feature hierarchy from deep semantic features to subtle fine-grained features, which consistently improves recognition accuracy for large to small-sized objects. As a result, MSPS encourages networks to generate richer representations of fine-grained objects and to be more flexible to scale changes.
\subsection{Visualization}
\label{subsec:vis}
To further investigate the proposed method, We present the visualization results in Fig.~\ref{fig5} and Fig.~\ref{fig6}.

Fig.~\ref{fig5} shows the selected patches at each stage by conducting MSPS for several images sampled from CUB. In each subfigure, the first column is the original image, and selected patches from stage-4 to stage-1 are marked with red rectangles. At higher stages, large-sized patches that capture several parts of the object are selected. Sufficiently large patches are appropriate for modeling the intra-image structure and overall semantics of a given object. On the other hand, at the lower stage, smaller patches are chosen to model subtle details. Especially, the smallest size patches selected in stage-1 capture the coarsest features such as object edges. As a result, MSPS enhances object representations at different levels for each stage.

Fig.~\ref{fig6} shows the cross-attention maps of MSCA for selected patches. For visualization, we sample some patches selected from stage-4. And then, we extract cross-attention maps between sampled patches and patches selected at different stages from SCA. In Fig.~\ref{fig6}, the first column shows the sampled stage-4 patch as a red rectangle. The second to fourth columns show the attention maps between sampled patches and selected patches from stage-3 to stage-1 as orange rectangles. The brightness of a color indicates the strength of attention: brighter means stronger interactions. In addition to modeling channel interactions in CCA, SCA models spatial interactions between selected multi-scale patches. As a result, each selected patch can be correlated with other patches that exist in different locations at different scales.
%
%
\section{Conclusion}
\label{sec:conclusion}
In this paper, we propose a multi-scale patch selection (MSPS) for fine-grained visual recognition (FGVR). MSPS selects salient patches in multi-scale at each stage of a multi-scale vision Transformer (MS-ViT) based on mean activation. In addition, we introduce class token transfer (CTT) and multi-scale cross-attention (MSCA) to effectively deal with selected multi-scale patch information. CTT transfers the globally updated \texttt{CLS} token to each stage so that stage-specific patch information can be shared throughout the entire network. MSCA directly models the spatial-/channel-wise correlation between selected multi-scale patches. Compared to single-scale patch selection (SSPS), MSPS provides richer representations for fine-grained objects and flexibility for scale changes. As a result, our proposed M2Former obtains accuracies of 92.4\%, and 91.1\% on CUB and NABirds, respectively, which outperform CNN-based models and ViT-based SSPS models. Our ablation experiments and further analyses validate the effectiveness of our proposed methods.
%
%

\end{document}